\pdfoutput=1
\documentclass[11pt,a4paper]{article}
\usepackage{style/acl2017}
\usepackage{times}
\usepackage{latexsym}
\usepackage{comment}
\usepackage{graphicx,xcolor}
\usepackage{url}

\usepackage{amsmath}
\usepackage{graphicx}
\usepackage{color}
\usepackage{amsfonts}

\newcommand{\vpLess}{\framebox[1.1\width]{$<$}}
\newcommand{\vpGreater}{\framebox[1.1\width]{$>$}}
\newcommand{\vpEqual}{\framebox[1.1\width]{$\simeq$}}

\aclfinalcopy 

\newcommand{\mf}[1]{{\color{blue}\{\textit{#1}\}$_{max}$}}

\title{\textsc{Verb Physics}: Relative Physical Knowledge of Actions and Objects}

\author{Maxwell Forbes \hspace{2em} Yejin Choi\\
 Paul G. Allen School of Computer Science \& Engineering \\
 University of Washington \\
  {\tt \{mbforbes,yejin\}@cs.washington.edu} \\}

\date{}

\begin{document}
\maketitle
\begin{abstract}

Learning commonsense knowledge from natural language text is nontrivial   
due to \emph{reporting bias}: people rarely state the obvious, e.g., ``My house is \emph{bigger} than me.''
However, while rarely stated explicitly, 
this trivial everyday knowledge 
does influence the way people talk about the world, which provides indirect clues to reason about the world. For example, a statement like, ``Tyler \emph{entered} his house'' 
implies that his house is \emph{bigger} than Tyler.

In this paper, we present an approach to infer relative physical knowledge of actions and objects along five dimensions (e.g., size, weight, and strength) from unstructured natural language text. 
We frame knowledge acquisition as joint inference over two closely related problems: learning (1) relative physical knowledge of object pairs and (2)  physical implications of actions when applied to those object pairs. 
Empirical results demonstrate that it is possible to extract knowledge of actions and objects from language and that joint inference over different types of knowledge improves performance.

\end{abstract}

\section{Introduction}
\label{sec:intro}

\begin{figure}[t!]

\begin{center}
\includegraphics[width=0.9\linewidth]{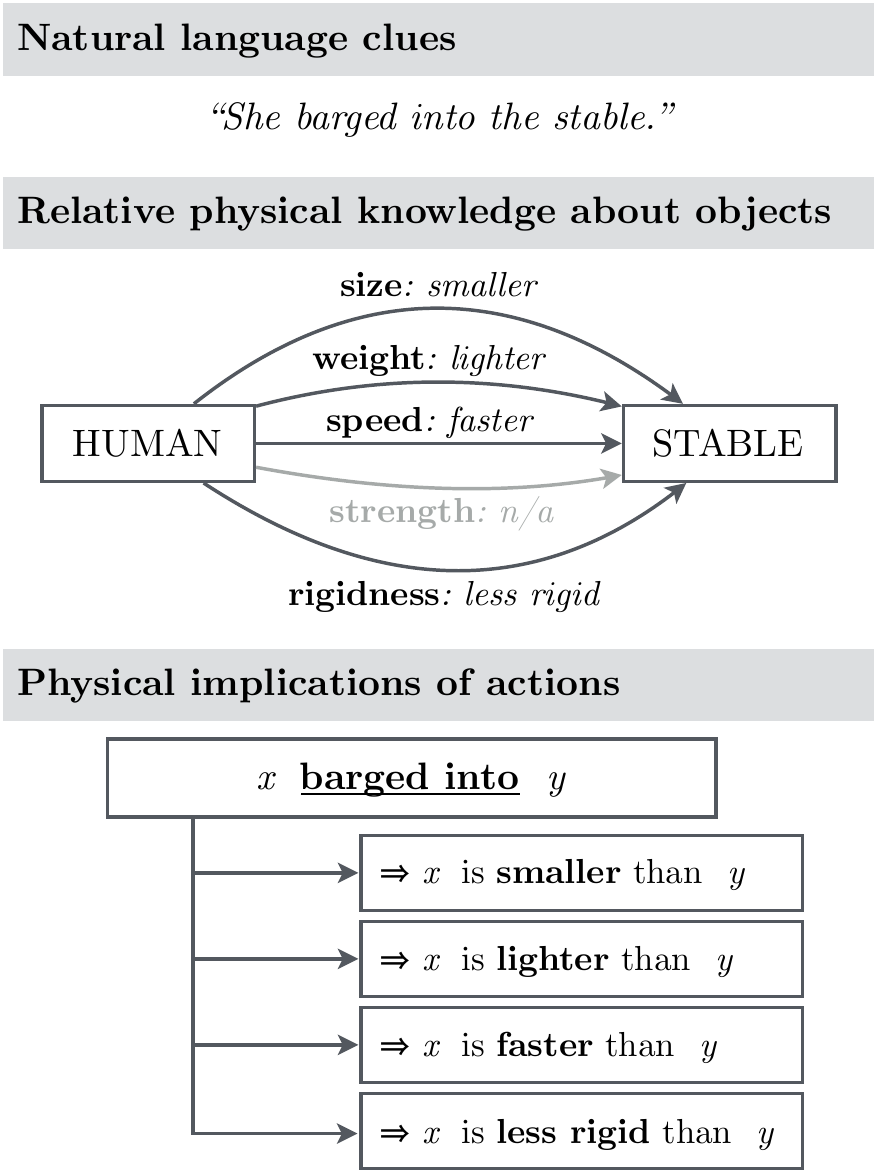}
\end{center}

\caption{An overview of our approach. A verb's usage in language (top) implies physical relations between objects it takes as arguments. This allows us to reason about properties of specific objects (middle), as well as the knowledge implied by the verb itself (bottom).}
\label{fig:frames-abstract}

\end{figure}

Reading and reasoning about natural language text often requires trivial  knowledge about everyday physical actions and objects. For example, given a sentence \textit{``Shanice could fit the trophy into the suitcase,''} we can trivially infer that the trophy must be smaller than the suitcase even though it is not stated explicitly. This reasoning requires knowledge about the action \emph{``fit''}---in particular, typical preconditions that need to be satisfied in order to perform the action. 
In addition, reasoning about the applicability of various physical actions in a given situation often requires background knowledge about objects in the world,  for example, that people are usually \emph{smaller} than houses, that cars generally move \emph{faster} than humans walk, or that a brick probably is \emph{heavier} than a feather. 

In fact, the potential use of such knowledge about everyday actions and objects can go beyond language understanding and reasoning. Many open challenges in computer vision and robotics may also benefit from such knowledge, as shown in recent work that requires visual reasoning and entailment \cite{izadinia2015segment,zhu2014reasoning}. Ideally, an AI system should acquire such knowledge through direct physical interactions with the world. However, such a physically interactive system does not seem feasible in the foreseeable future. 

In this paper, we present an approach to acquire trivial physical knowledge from unstructured natural language text as an alternative knowledge source. In particular, we focus on acquiring relative physical knowledge of actions and objects organized along five dimensions: size, weight, strength, rigidness, and speed. Figure~\ref{fig:frames-abstract} illustrates example knowledge of (1) relative physical relations of object pairs and (2) physical implications of actions when applied to those object pairs.  

While natural language text is a rich source to obtain broad knowledge about the world, compiling trivial commonsense knowledge from unstructured text is a nontrivial feat.
The central challenge lies in \emph{reporting bias}: people rarely states the obvious~\cite{van2010extracting,sorower2011inverting,gordon2013reporting,misra2016seeing,zhang2017ordinal}, since it goes against Grice's conversational maxim on the quantity of information \cite{grice1975logic}.

In this work, we demonstrate that it is possible to overcome reporting bias and still extract the unspoken knowledge from language. The key insight is this: there is consistency in the way people describe how they interact with the world, which provides vital clues to reverse engineer the common knowledge shared among people. More concretely, we frame knowledge acquisition as joint inference over two closely related puzzles: inferring relative physical knowledge about object pairs while simultaneously reasoning about physical implications of actions. 

Importantly, four of five dimensions of knowledge in our study---weight, strength, rigidness, and speed---are either not visual or not easily recognizable by image recognition using currently available computer vision techniques. Thus, our work provides unique value to complement recent attempts to acquire commonsense knowledge from web images \cite{izadinia2015segment,bagherinezhad2016elephants,sadeghi2015viske}.

In sum, our contributions are threefold:

\vspace*{-1mm}
\begin{itemize}

	\item 
	We introduce a new task in the domain of commonsense knowledge extraction from language, focusing on the physical implications of actions and the relative physical relations among objects, organized along five dimensions. 
	\vspace*{-2mm}

	\item 
	We propose a model that can infer relations over grounded object pairs together with first order relations implied by physical verbs.
	\vspace*{-2mm}

	\item 
	We develop a new dataset \textsc{VerbPhysics} that compiles crowdsourced knowledge of actions and objects.\footnote{\url{https://uwnlp.github.io/verbphysics/}}

\end{itemize}
\vspace*{-1mm}

\noindent
The rest of the paper is organized as follows. We first provide the formal definition of knowledge we aim to learn in Section~\ref{sec:approach}. We then describe our data collection in Section~\ref{sec:data} and present our inference model in Section~\ref{sec:model}. Empirical results are given in Section~\ref{sec:experiments} and discussed in Section~\ref{sec:discussion}. We review related work in Section~\ref{sec:relatedwork} and conclude in Section~\ref{sec:conclusion}.
\section{Representation of Relative Physical Knowledge}
\label{sec:approach}

\begin{figure*}

\begin{center}
\includegraphics[width=0.95\textwidth]{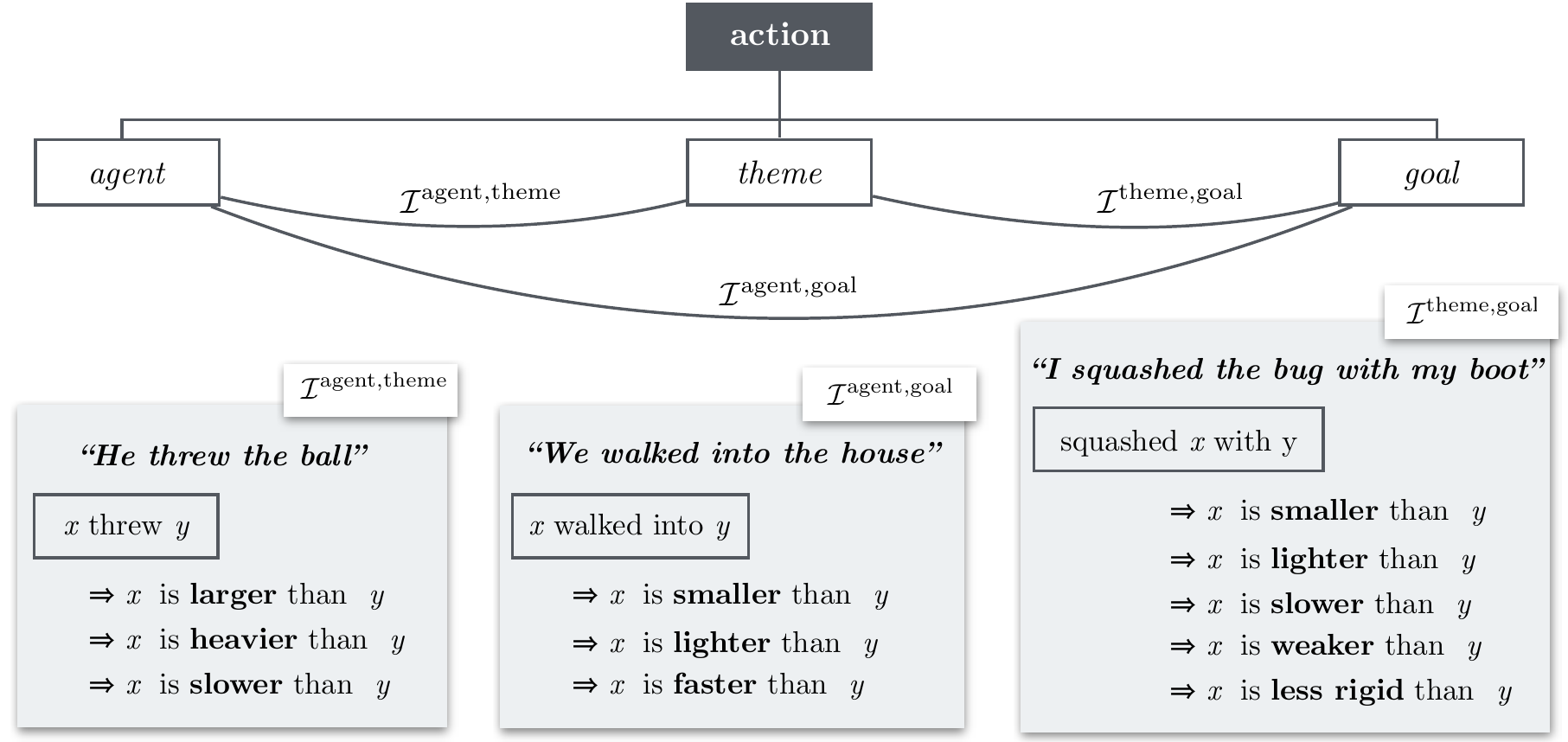}

\caption{
Example physical implications represented as frame relations between a pair of arguments.
}
\label{fig:frame-examples}
\end{center}
\end{figure*}

\subsection{Knowledge Dimensions}
We consider five dimensions of relative physical knowledge in this work: \emph{size, weight, strength,  rigidness,}  and  \emph{speed.} ``Strength'' in our work refers to the physical durability of an object (e.g., ``diamond'' is stronger than ``glass''), while ``rigidness'' refers to the physical flexibility of an object (e.g., ``glass'' is more rigid than a ``wire''). 
When considered in verb implications, \emph{size, weight, strength,} and \emph{rigidness} concern individual-level semantics; the relative properties implied by verbs in these dimensions are true in general. On the other hand, \emph{speed} concerns stage-level semantics; its implied relations hold only during a window surrounding the verb \cite{carlson1977unified}.

\subsection{Relative physical knowledge}

Let us first consider the problem of representing  
relative physical knowledge
between two objects. We can write a single piece of knowledge like ``A person is larger than a basketball'' as

$$
\texttt{person} >^\mathrm{size} \texttt{basketball}
$$

\noindent
Any propositional statement can have exceptions and counterexamples. Moreover, we need to cope with uncertainties involved in knowledge acquisition. Therefore, we assume each piece of knowledge is associated with a probability distribution.
More formally, given objects $x$ and $y$, we define a random variable $O_{x,y}^{a}$ whose range is $\{\vpGreater{}, \vpLess{}, \vpEqual{}\}$ with respect to a knowledge dimension $a \in \{$\textsc{size},\textsc{weight},\textsc{strength},\textsc{rigidness},\textsc{speed}$\}$ so that: 
$$
\mathbb{P}(O_{x,y}^a = r), r \in \{\vpGreater{}, \vpLess{}, \vpEqual{}\}.
$$

\noindent
This immediately provides two simple properties:

$$
\begin{aligned}[t]
\mathbb{P}(O_{x,y} = \vpGreater{}) &= \mathbb{P}(O_{y,x} = \vpLess{})\\
\mathbb{P}(O_{x,x} = \vpEqual{}) &= 1\\
\end{aligned}
$$

\subsection{Physical Implications of Verbs}

Next we consider representing 
relative physical implications of actions applied over 
two objects. For example, consider an action frame ``$x$ threw
$y$.'' 
In general, following implications are likely to be true: 

$$
\begin{aligned}[t]
\text{``}x \text{ threw } y\text{''} &\implies x >^\mathrm{size} y\\
\text{``}x \text{ threw } y\text{''} &\implies x >^\mathrm{weight} y\\
\text{``}x \text{ threw } y\text{''} &\implies x <^\mathrm{speed} y\\
\end{aligned}
$$

\noindent
Again, in order to cope with exceptions and uncertainties, we assume a probability distribution associated with each implication.
More formally, we define
a random variable $F_v^a$ to denote the implication of the action verb $v$ when applied over its arguments $x$ and $y$ with respect to a knowledge dimension $a$ so that:

\begin{align*}\small
& \mathbb{P}(F_\mathrm{threw}^\mathrm{size} = \vpGreater{}) 
   := \mathbb{P}(\text{``}x \text{ threw } y\text{''} \Rightarrow x >^\mathrm{size} y) \\
& \mathbb{P}(F_\mathrm{threw}^\mathrm{wgt} = \vpGreater{}) 
   := \mathbb{P}(\text{``}x \text{ threw } y\text{''} \Rightarrow x >^\mathrm{wgt} y)
\end{align*}

\noindent
where the range of $F_{threw}^{size}$ is $\{\vpGreater{}, \vpLess{}, \vpEqual{}\}$.
Intuitively, $F_{threw}^{size}$ represents the likely first order relation implied by ``throw'' over ungrounded (i.e., variable) object pairs.

The above definition assumes that there is only a single implication relation for any given verb with respect to a specific knowledge dimension. This is generally not true, since a verb, especially a common action verb, can often invoke a number of different frames according to frame semantics \cite{fillmore1976frame}. 
Thus, given a number of different frame relations $v_1 ... v_T$ associated with a verb $v$, we define random variables $F$ with respect to a specific frame relation $v_t$, i.e., $F_{v_t}^a$. We use this notation going forward.

\paragraph{Frame Perspective on Verb Implications:}
Figure~\ref{fig:frame-examples} illustrates the frame-centric view of physical implication knowledge we aim to learn. 
Importantly, the key insight of our work is inspired by Fillmore's original manuscript on frame semantics~\cite{fillmore1976frame}.
Fillmore has argued that ``frames''---the contexts in
which utterances are situated---should be considered as a third primitive of
describing a language, along with a grammar and lexicon. 
While existing frame annotations such as FrameNet
\cite{baker1998berkeley}, PropBank
\cite{palmer2005proposition}, and VerbNet \cite{kipper2000class} provide rich frame knowledge associated with a predicate, 
none of them provide the exact kind of physical implications we consider in our paper.
Thus, our work can potentially contribute to these resources by investigating new approaches to automatically recover richer frame knowledge from language.
In addition, our work is motivated by the formal semantics of \citet{dowty1991thematic}, as the task of learning verb implications is essentially that of extracting lexical entailments for verbs.

\section{Data and Crowdsourced Knowledge}
\label{sec:data}

\paragraph{Action Verbs:} We pick 50 classes of Levin verbs from both
``alternation classes'' and ``verb classes'' \cite{levin1993english}, which
corresponds to about 1100 unique verbs. We sort this list by frequency of
occurrence in our frame patterns in the Google Syntax Ngrams corpus
\cite{goldberg2013dataset} and pick the top 100 verbs.

\paragraph{Action Frames:} 

Figure~\ref{fig:frame-examples} illustrates examples of action frame relations. 
Because we consider implications over pairwise argument relations for each frame, there are sometimes multiple frame relations we consider for a single frame. 
To enumerate action frame relations for each verb, we use syntactic patterns based on dependency parse by extracting the core components (subject, verb, direct object, prepositional object) of an action, then map the subject to an agent, the direct object to a theme, and the prepositional object to a goal.\footnote{Future research could use an SRL parser instead. We use dependency parse to benefit from the Google Syntax Ngram dataset that provides language statistics over an extremely large corpus, which does not exist for SRL.} 
For those frames that involve an argument in a prepositional phrase, we create a separate frame for each preposition based on the statistics observed in the Google Syntax Ngram corpus.
%

Because the syntax ngram corpus provides only tree snippets without context, this way of enumerating potential frame patterns tend to over-generate.
Thus we refine our prepositions for each frame by taking either the intersection or union with the top 5 Google Surface Ngrams \cite{michel2011quantitative},
depending on whether the frame was under- or over-generating. We also add an
additional crowdsourcing step where we ask crowd workers to judge whether a
frame pattern with a particular verb and preposition could plausibly be found in
a sentence. This process results in 813 frame templates, an average of 8.13 per verb.

\paragraph{Object Pairs:} To provide a source of ground truth relations between objects, we select the object pairs that occur in the 813 frame templates with positive pointwise mutual information (PMI) across the Google Syntax Ngram corpus. After replacing a small set of ``human'' nouns with a generic \textsc{Human} object, filtering out nouns
labeled as abstract by WordNet \cite{miller1995wordnet}, and distilling all surface forms to their lemmas (also with WordNet), the result is 3656 object pairs.

\subsection{Crowdsourcing Knowledge}

%

\begin{table}
\footnotesize

%
%

\begin{center}
\begin{tabular}{ l|c|c }
 \hline
 \multicolumn{3}{|c|}{Data collected} \\
 \hline
       & \textbf{Total} & \textbf{Seed / dev / test} \\
 \hline
   Verbs$_{5\%}$ & 100 & 5 / 45 / 50 \\
   Verbs$_{20\%}$ & '' & 20 / 30 / 50 \\   
 Frames$_{5\%}$ & 813 & 65 / 333 / 415 \\
 Frames$_{20\%}$ & '' & 188 / 210 / 415 \\ 
 Object pairs$_{5\%}$ & 3656 & 183 / 1645 / 1828 \\
 Object pairs$_{20\%}$ & '' & 733 / 1096 / 1828 \\ 
\end{tabular}
\end{center}

\vspace{0.3em}

%
%

\begin{center}

\begin{tabular}{ l|c|c|c|c }
 \hline
 \multicolumn{5}{|c|}{Per attribute frame statistics} \\
 \hline
        & \multicolumn{2}{|c|}{\textit{Agreement}} & \multicolumn{2}{|c}{\textit{Counts (usable)}} \\
 \cline{2-5}
 & \textbf{2/3} & \textbf{3/3} & \textbf{Verbs} & \textbf{Frames} \\
 \hline
 size & 0.91 & 0.41 & 96 & 615 \\
 weight & 0.90 & 0.33 & 97 & 562 \\
 strength & 0.88 & 0.25 & 95 & 465 \\
 rigidness & 0.87 & 0.26 & 89 & 432 \\
 speed & 0.93 & 0.36 & 88 & 420 \\
\end{tabular}
\end{center}

\vspace{0.3em}

%
%

\begin{center}

\begin{tabular}{ l|c|c|c|c }
 \hline
 \multicolumn{5}{|c|}{Per attribute object pair statistics} \\
 \hline
        & \multicolumn{2}{|c|}{\textit{Agreement}} & \multicolumn{2}{|c}{\textit{Counts (usable)}} \\
 \cline{2-5}
 & \textbf{2/3} & \textbf{3/3} & \textbf{Distinct objs} & \textbf{Pairs} \\
 \hline
size       & 0.95 & 0.59 & 210 & 2552 \\
weight     & 0.95 & 0.56 & 212 & 2586 \\
strength   & 0.92 & 0.43 & 208 & 2335 \\
rigidness  & 0.91 & 0.39 & 212 & 2355 \\
speed      & 0.90 & 0.38 & 209 & 2184 \\
\end{tabular}
\end{center}

\caption{Statistics of crowdsourced knowledge. Frames are partitioned by verb. Counts are shown for \textit{usable} data, which includes only $\geq$ 2/3 agreement and removes all with ``no relation.'' Each prediction task (frames or object pairs) is given 5\% of that domain's data as seed. We compare models using either 5\% or 20\% of the \emph{other} domain's data as seed.
}

\label{table:data-stats}

\end{table}

We collect human judgements of the frame knowledge implications to use as a small set of seed knowledge (5\%), a development set (45\%), and a test set (50\%). 
Crowd workers are given with a frame template such as ``\textbf{x} threw
\textbf{y},'' and then asked to list a few plausible objects (including people
and animals) for the missing slots (e.g., \textbf{x} and
\textbf{y}).\footnote{This step is to prime them for thinking about the
particular template; we do not use the objects they provided.} We then ask
them to rate the general relationship that the arguments of the frame exhibit
with respect to all knowledge dimensions (size, weight, etc.). For each knowledge dimension, or attribute, $a$, workers select an answer 
from (1) $\mathbf{x} >^a \mathbf{y}$, (2) $\mathbf{x} <^a \mathbf{y}$,
(3) $\mathbf{x} \simeq^a \mathbf{y}$, or (4) no general relation.

We conduct a similar crowdsourcing step for the set of object pairs. We ask crowd workers to compare each of the 3656 object pairs along the five knowledge dimensions we consider, selecting an answer from the same options above as with frames. We reserve 50\% of the data as a test set, and split the remainder up either 5\% / 45\% or 20\% / 30\% (seed / development) to investigate the effects of different seed knowledge sizes on the model.

Statistics for the dataset are provided in Table~\ref{table:data-stats}. About
90\% of the frames as well as object pairs had 2/3 agreement between workers. After removing
frame/attribute combinations and object pairs that received less than 2/3 agreement, or were
selected by at least 2/3 workers to have no relation, we end up with roughly
400--600 usable frames and 2100--2500 usable object pairs per attribute.
\section{Model}
\label{sec:model}

\begin{figure*}

\includegraphics[width=\textwidth]{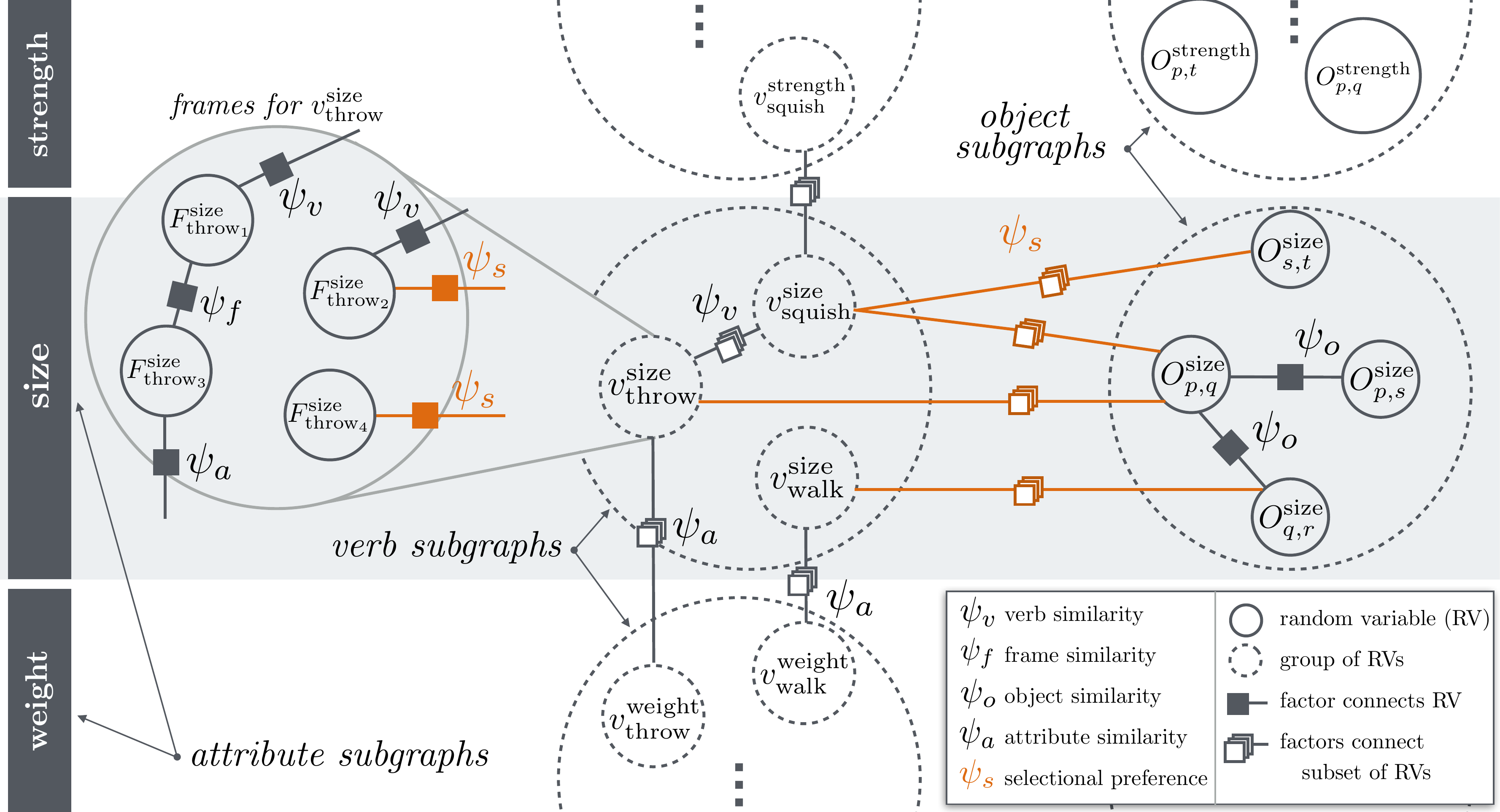}

\caption{High level view of the factor graph model. Performance on both learning
relative knowledge about objects (right), as well as entailed knowledge from
verbs (center) via realized frames (left), is improved by modeling their
interplay (orange). Unary seed ($\psi_{seed}$) and embedding ($\psi_{emb}$) factors
are omitted for clarity.}
\label{fig:model}

\end{figure*}

We model knowledge acquisition as probabilistic inference over a factor graph of knowledge. As shown in Figure~\ref{fig:model}, the graph consists of multiple substrates (page-wide boxes) corresponding to different knowledge dimensions (shown only three of them ---strength, size, weight---for brevity).
Each substrate consists of two types of sub-graphs: verb subgraphs and object subgraphs, which are connected through factors that quantify action--object compatibilities. Connecting across substrates are factors that model inter-dependencies across different knowledge dimensions. In what follows, we describe each graph component.

\subsection{Nodes}

The factor graph contains two types of nodes in order to capture two classes
of knowledge. The first type of nodes are object pair nodes. Each object
pair node is a random variable $O_{x,y}^a$ which captures the relative strength
of an attribute $a$ between objects $x$ and $y$.

The second type of nodes are frame nodes. Each frame node is a random variable
$F_{v_t}^a$. This corresponds to the verb $v$ used in a particular type of frame
$t$, and captures the implied knowledge the frame $v_t$ holds along an attribute
$a$.

All random variables take on the values $\{\vpGreater{}, \vpLess{}, \vpEqual{}\}$. For an object pair
node $O_{x,y}^a$, the value represents the belief about the relation between $x$
and $y$ along the attribute $a$. For a frame node $F_{v_t}^a$, the value represents
the belief about the relation along the attribute $a$ between \textit{any} two
objects that might be used in the frame $v_t$.

We denote the sets of all object pair and frame random variables $\mathcal{O}$
and $\mathcal{F}$, respectively.

\subsection{Action--Object Compatibility}
The key aspect of our work is to reason about two types of knowledge simultaneously: relative knowledge of grounded object pairs, and implications of actions related to those objects. 
Thus we connect the verb subgraphs and object subgraphs through selectional preference factors $\psi_
{s}$ between two such nodes $O_{x,y}^a$ and $F_{v_t}^a$ if we find
evidence from text that suggests objects $x$ and $y$ are used in the frame
$v_t$. These factors encourage both random variables to agree on the same value.

As an example, consider a node $O_{p,b}^{size}$ which represents the relative
size of a person and a basketball, and a node $F_{threw_{dobj}}^{size}$ which
represents the relative size implied by an \textit{``$x$ threw $y$''} frame. If
we find significant evidence in text that \textit{``[person] threw
[basketball]''} occurs, we would add a selectional preference factor to connect
$O_{p,b}^{size}$ with $F_{threw_{dobj}}^{size}$ and encourage them towards the
same value. This means that if it is discovered that people are larger than
basketballs (the value $\vpGreater{}$), then we would expect the frame \textit{``$x$ threw $y$''}
to entail $x >^{size} y$ (also the value $\vpGreater{}$).

\subsection{Semantic Similarities}

Some frames have relatively sparse text evidences to support their corresponding knowledge acquisition. Thus, we include several types of factors based on semantic similarities as described below. 

\paragraph{Cross-Verb Frame Similarity:} 
We add a group of factors $\psi_{v}$ between two verbs $v$ and $u$ (to connect a specific frame of $v$ with a corresponding frame of $u$) based on the verb-level similarities.

\paragraph{Within-Verb Frame Similarity:} 
Within each verb $v$, which consists of a set of frame relations $v_1,...v_T$, we also include frame-level similarity factors $\psi_{f}$ between 
$v_i$ and $v_j$. This gives us more evidence
over a broader range of frames when textual evidence might be sparse.

\paragraph{Object Similarity:} As with verbs, we add factors $\psi_{o}$
that encourage similar pairs of objects to take the same value. Given that each
node represents a pair of objects, finding that $x$ and $y$ are similar yields
two main cases in how to add factors (aside from the trivial case where the
variable $O_{x,y}^a$ is given a unary factor to encourage the value $\vpEqual{}$).

\begin{enumerate}
	\item If nodes $O_{x,z}$ and $O_{y,z}$ exist, we expect objects $x$ and $y$
	to both have a similar relation to $z$. We add a factor that encourages
	$O_{x,z}$ and $O_{y,z}$ to take the same value. The same is true if nodes
	$O_{z,x}$ and $O_{z,y}$ exist.

    \vspace*{-2mm}
	\item On the other hand, if nodes $O_{x,z}$ and $O_{z,y}$ exist, we expect
	these two nodes to reach the opposite decision. In this case, we add a
	factor that encourages one node to take the value \texttt{$\vpGreater{}$} if the other
	prefers the value \texttt{$\vpLess{}$}, and vice versa. (For the case of
	\texttt{$\vpEqual{}$}, if one prefers that value, then both should.)
\end{enumerate}

\subsection{Cross-Knowledge Correlation} 
Some knowledge dimensions, such as size and weight, have
a significant correlation in their implied relations. For two such
attributes $a$ and $b$, if the same frame $F^a_{v_i}$ and $F^b_{v_i}$ exists in
both graphs, we add a factor $\psi_{a}$ between them to push them towards
taking the same value.

\begin{table*}\small

\begin{center}
\begin{tabular}{ c||ccccc|c||ccccc|c }

\hline

                   & \multicolumn{6}{c||}{\bf Development} & \multicolumn{6}{c}{\bf Test} \\

\cline{2-13}

\textbf{Algorithm} & size & weight & stren & rigid & speed & \textit{overall} & size & weight & stren & rigid & speed & \textit{overall} \\

\hline

\textsc{random} & 0.33 & 0.33 &  0.33 &  0.33 &  0.33 &  0.33 &  0.33 &  0.33 &  0.33 &  0.33 &  0.33 &  0.33 \\

\textsc{majority} & 0.38 & 0.41 & 0.42 & 0.18 & \textbf{0.83} & 0.43 & 0.35 & 0.35 & 0.43 & 0.20 & 0.88 & 0.44 \\

\textsc{emb-maxent} & 0.62 & 0.64 & 0.60 & \textbf{0.83} & \textbf{0.83} & 0.69 & 0.55 & 0.55 & 0.59 & 0.79 & 0.88 & 0.66 \\

\textsc{our model (a)}& 0.71 & 0.63 & 0.61 & 0.82 & \textbf{0.83} & 0.71 & 0.55 & 0.55 & 0.55 & 0.79 & \textbf{0.89} & 0.65  \\

\textsc{our model (b)} & \textbf{0.75} & \textbf{0.68} & \textbf{0.68} & 0.82 & 0.78 & \textbf{0.74} & \textbf{0.74} & \textbf{0.71} & \textbf{0.65} & \textbf{0.80} & 0.87 & \textbf{0.75} \\

\hline

\end{tabular}
\end{center}

\begin{center}
\begin{tabular}{ c||ccccc|c||ccccc|c }

\hline

                   & \multicolumn{6}{c||}{\bf Development} & \multicolumn{6}{c}{\bf Test} \\

\cline{2-13}

\textbf{Algorithm} & size & weight & stren & rigid & speed & \textit{overall} & size & weight & stren & rigid & speed & \textit{overall} \\

\hline

\textsc{random} & 0.33 & 0.33 &  0.33 &  0.33 &  0.33 &  0.33 &  0.33 &  0.33 &  0.33 &  0.33 &  0.33 &  0.33 \\

\textsc{majority} & 0.50 &  0.54 &  0.51 &  0.50 &  0.53 &  0.51 &  0.51 &  0.55 &  0.52 &  0.49 &  0.50 &  0.51  \\


\textsc{emb-maxent} & 0.68 & 0.66 & 0.64 & 0.67 & 0.65 & 0.66 & 0.71 & 0.67 & 0.64 & 0.65  & \textbf{0.63}  & 0.66  \\

\textsc{our model (a)} & 0.74 & 0.69 & 0.67 & \textbf{0.68} & \textbf{0.66} & 0.69 & 0.68 & 0.70 & 0.66 & \textbf{0.66} & 0.60 & 0.66 \\

\textsc{our model (b)} & \textbf{0.75} & \textbf{0.74} & \textbf{0.71} & \textbf{0.68} & \textbf{0.66} & \textbf{0.71}               & \textbf{0.75} & \textbf{0.76} & \textbf{0.72} & 0.65 & 0.61 & \textbf{0.70} \\

\hline

\end{tabular}
\end{center}

\caption{Accuracy of baselines and our model on both tasks. Top: frame prediction task; bottom: object pair prediction task. In both tasks 5\% of in-domain data (frames or object pairs, respectively) are available as seed data. We compare providing the other type of data (object pairs or frames, respectively) as seed knowledge, trying 5\% (\textsc{our model (a)}) and 20\% (\textsc{our model (b)}).}

\label{table:resultsframepred}

\end{table*}

\subsection{Seed Knowledge} In order to kick off learning, 
we provide a small set of seed knowledge among 
the random variables in $\{\mathcal{O}, \mathcal{F}\}$ with seed factors $\psi_{seed}$. 
These unary seed factors
push the belief for its associated random variable strongly towards the seed label.

\subsection{Potential Functions}

\label{sec:pot-funcs}

\paragraph{Unary Factors:} For all frame and object pair random variables in the training set, we train a maximum entropy classifier to predict the value of the variable. We then use the probabilities of the classifier as potentials for seed factors given to all random variables in their class (frame or object pair).
Each log-linear classifier is trained separately per attribute on a featurized vector of the variable:
$$
\mathbb{P}(r|X^a) \propto e^{w_a \cdot f(X^a)}
$$
\noindent
The feature function is defined differently according to the node type:
\begin{align*}
    f(O^a_{p,q}) &:= \langle g(p), g(q) \rangle \\
    f(F^a_{v_t}) &:= \langle h(t), g(v), g(t) \rangle   
\end{align*}
\noindent
Here $g(x)$ is the GloVe word embedding \cite{pennington2014glove} for the word $x$ ($t$ is the frame relation's preposition, and $g(t)$ is simply set to the zero vector if there is no preposition) and $h(t)$ is a one-hot vector of the frame relation type. We use GloVe vectors of 100 dimensions for verbs and 50 dimensions for objects and prepositions (the dimensions picked based on development set).

\paragraph{Binary Factors:} 
In the case of all other factors, 
we use a ``soft 1'' agreement matrix with strong signal down the diagonals:
\begin{align*}\small
\begin{bmatrix}
    & > & \simeq & <\\
    > & {\bf 0.7} & 0.1 & 0.2 \\
    \simeq & 0.15 & {\bf 0.7} & 0.15 \\    
    < & 0.2 & 0.1 & {\bf 0.7}
\end{bmatrix}    
\end{align*}
\noindent

\subsection{Inference} After our full graph is constructed, we use belief propagation to infer the assignments of frames and object pairs not in our training data. Each message $\mu$ is a vector where each element is the probability that a random variable takes on each value $x \in \{\vpGreater{}, \vpLess{}, \vpEqual\}$. A message passed from a random variable $v$ to a neighboring factor $f$ about the value $x$ is the product of the messages from its other neighboring factors about $x$:

$$
\mu_{v \rightarrow f}(x) \propto \prod_{f' \in N(v) \backslash \{ f \}} \mu_{f' \rightarrow v}(x)
$$

\noindent
A message passed from a factor $f$ with potential $\psi$ to a random variable $v$ about its value $x$ is a marginalized belief about $v$ taking value $x$ from the other neighboring random variables combined with its potential:
$$
\mu_{f \rightarrow v}(x) \propto \sum_{\mathbf{x}:\mathbf{x}[v]=x} \psi(\mathbf{x}) \prod_{v' \in N(f) \backslash \{ v \} } \mu_{v' \rightarrow f}(\mathbf{x}[v'])
$$

After stopping belief propagation, the marginals for a node can be computed and used as
a decision for that random variable. The marginal for $v$ taking value $x$ is the product of its surrounding factors' messages:

$$
v(x) \propto \prod_{f \in N(v)} \mu_{f \rightarrow v}(x)
$$
\section{Experimental Results}

\label{sec:experiments}

\paragraph{Factor Graph Construction:} We first need to pick a set of frames and
objects to determine our set of random variables. The frames are simply the
subset of the frames that were crowdsourced in the given configuration (e.g.,
seed + dev), with ``soft 1'' unary seed factors (the gold label indexed row of the binary factor matrix) given only to those in the seed set. The same selection criteria and seed factors are applied to the crowdsourced object pairs.

For lexical similarity factors ($\psi_v$, $\psi_o$), we pick connections based
on the cosine similarity scores of GloVe vectors thresholded above a value chosen
based on development set performance.
Attribute similarity factors ($\psi_a$) are chosen based on sets of frames that
reach largely the same decisions on the seed data (95\%). Frame similarity
factors ($\psi_f$) are added to pairs of frames with linguistically similar
constructions. Finally, selectional preference factors ($\psi_s$) are picked by using a 
threshold (also tuned on the development set) of pointwise mutual information (PMI) between the frames and the object pairs' occurrences in the Google Syntax Ngram corpus.

For each task, we consider the set of factors to include in each model a hyperparameter, which is also tuned on the development set.

\paragraph{Baselines:} Baselines include making a \textsc{random} choice, 
picking between \vpGreater{}, \vpLess{}, and \vpEqual{}), 
picking the \textsc{majority} label, and a maximum entropy classifier based on the embedding representations (\textsc{emb-maxent}) defined in Section~\ref{sec:pot-funcs}.

\paragraph{Inferring Knowledge of Actions:}

\begin{figure}[t!]

\begin{center}
\includegraphics[width=0.99\linewidth]{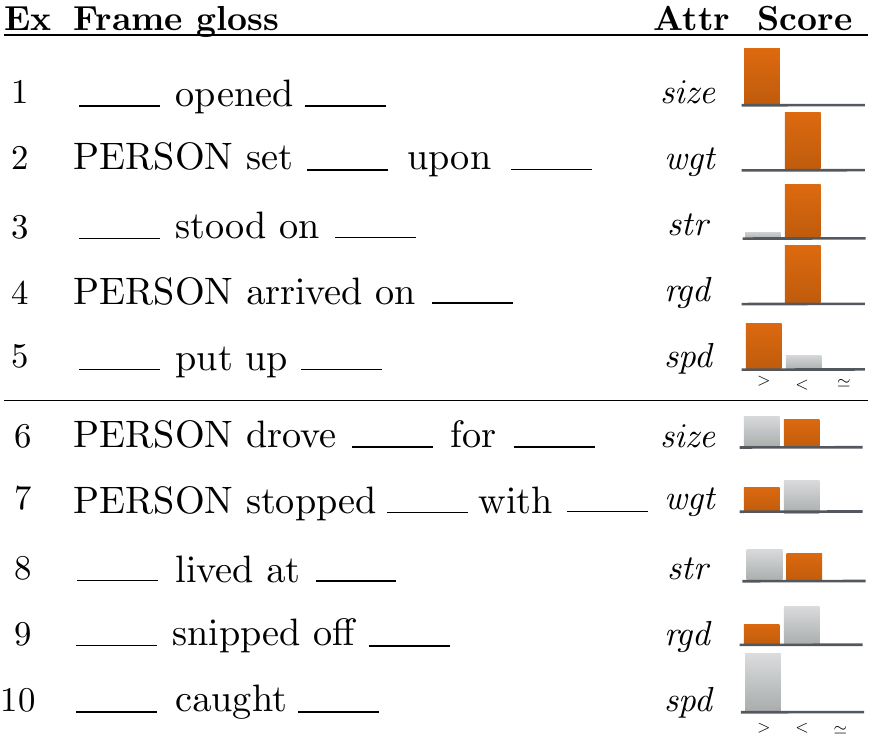}
\end{center}

\caption{Example model predictions on dev set frames. The model's confidence is shown by the bars on the right. The correct relation is highlighted in orange (6--10 are failure cases for the model). If there are two blanks, the relation is between them. If there is only one blank, the relation is between \textsc{Person} and the blank. Note that \vpEqual{} receives minuscule weight because it is never the correct value for frames in the seed set.}
\label{fig:results-examples-frames}

\end{figure}

Our first experiment is to predict knowledge implied by new frames. In this task, 5\% of the frames are available as seed knowledge.
We experiment with two different sets of seed knowledge for the object pair data: \textsc{our model (a)} uses only 5\% of the object pair data as seed, and \textsc{our model (b)} uses 20\%.

The full results for the baseline methods and our model are given in the upper half of Table~\ref{table:resultsframepred}. 
Our model outperforms the baselines on all attributes except for the speed, which has a highly skewed label distribution to allow the majority baseline to perform well. 
Ablations are given in Table~\ref{table:ablations}, and sample correct predictions from the development set are shown in examples 1--5 of Figure~\ref{fig:results-examples-frames}.

\begin{table}\small
\begin{center}
\begin{tabular}{ l|c }
 \textbf{Ablated (or added) component} & \textbf{Accuracy} \\
 \hline
\: \: -- Verb similarity & 0.69 \\
\: \:  + Frame similarity & 0.62 \\
\: \: -- Action-object compatibility & 0.62 \\
\: \: -- Object similarity & 0.70 \\
\: \:  + Attribute similarity & 0.62 \\
\: \: -- Frame embeddings & 0.63 \\
\: \: -- Frame seeds & 0.62 \\
\: \: -- Object embeddings & 0.62 \\
\: \: -- Object seeds & 0.62 \\
 \hline
 \hline
  \textsc{our model (a)} & \textbf{0.71} \\
\end{tabular}
\end{center}

\caption{Ablation results on \emph{size} attribute for the frame prediction task on the development dataset for \textsc{our model (a)} (5\% of the object pairs as seed data). We find that different graph configurations improve performance for different tasks and data amounts. In this setting, frame and attribute similarity factors hindered performance.}

\label{table:ablations}

\end{table}

\paragraph{Inferring Knowledge of Objects:} Our second experiment is to predict the correct relations of new object pairs. The data for this task is the inverse of before: 5\% of the object pairs are available as seed knowledge, and we experiment with both 5\% (\textsc{our model (a)}) and 20\% (\textsc{our model (b)}) frames given as seed data. Again, both are independently tuned on the development data. Results for this task are presented in the lower half of Table~\ref{table:resultsframepred}. While \textsc{our model (a)} is competitive with the strongest baseline, introducing the additional frame data allows \textsc{our model (b)} to reach the highest accuracy.
\section{Discussion}

\label{sec:discussion}

\paragraph{Metaphorical Language:} While our frame patterns are intended to capture action verbs, our templates also match senses of those verbs that can be used with abstract or metaphorical arguments, rather than directly physical ones. One example from the development set is ``$x$ contained $y$.'' While $x$ and $y$ can be real objects, more abstract senses of ``contained'' could involve $y$ as a ``forest fire'' or even a ``revolution.'' In these instances, $x >^\mathrm{size} y$ is plausible as an abstract notion: if some entity can contain a revolution, we might think that entity as ``larger'' or ``stronger'' than the revolution. 

\paragraph{Error analysis:} Examples 6--10 in Figure~\ref{fig:results-examples-frames} highlight failure cases for the model. Example 6 shows a case where the comparison is nonsensical because \textit{``for''} would naturally be followed by a purpose (\textit{``He drove the car for work.''}) or a duration (\textit{``She drove the car for hours.''}) rather than a concrete object whose size is measurable.
Example 7 highlights an underspecified frame. One crowd worker provided the example, \textit{``\textsc{Person} stopped the fly with \{the jar / a swatter\},''} where $\mathrm{fly} <^\mathrm{weight} \{\mathrm{jar, swatter}\}$. However, two crowd workers provided examples like \textit{``\textsc{Person} stopped their car with the brake,''} where clearly $\mathrm{car} >^\mathrm{weight} \mathrm{brake}$. This example illustrates complex underlying physics we do not model: a brake---the pedal itself---is used to stop a car, but it does so by applying significant force through a separate system.

The next two examples are cases where the model nearly predicts correctly (8, e.g., \textit{``She lived at the office.''}) and is just clearly wrong (9, e.g., \textit{``He snipped off a locket of hair''}). Example 10 demonstrates a case of polysemy where the model picks the wrong side. In the phrase, \textit{``She caught the runner in first,''}, it is correct that $\mathrm{she} >^\mathrm{speed} \mathrm{runner}$. However, the sense chosen by the crowd workers is that of, \textit{``She caught the baseball,''} where indeed $\mathrm{she} <^\mathrm{speed} \mathrm{baseball}$.
\section{Related work}

\label{sec:relatedwork}

%
%

%
%

Several works straddle the gap between IE, knowledge base completion, and
learning commonsense knowledge from text. Earlier works in these areas use large
amounts of text to try to extract general statements like ``\textsc{a thing can
be readable}'' \cite{gordon2010learning} and frequencies of events
\cite{gordon2012using}. Our work focuses on specific domains of knowledge
rather than general statements or occurrence statistics, and develops a frame-centric approach to circumvent reporting bias. Other work uses a knowledge base and scores unseen tuples based on similarity to existing ones
\cite{angeli2013philosophers,li2016commonsense}, or extends it by inferring new facts from unstructured text using natural language inference \cite{angeli2014naturalli}. \citet{zhang2017ordinal} predict the likelihood of entailed commonsense statements extracted from a large text corpus. 
In contrast to the above, our work seeks to induce several novel types of graded physical knowledge which lack existing databases.

A number of recent works combine multimodal input to learn visual attributes \cite{bruni2012distributional,silberer2013models}, extract commonsense knowledge from web images \cite{tandon2016commonsense}, and overcome reporting bias \cite{misra2016seeing}.
In contrast, we focus on natural language evidence to reason about attributes that are both in (size) and out (weight, rigidness, etc.) of the scope of computer vision.
Yet other works mine numerical attributes of objects \cite{narisawa2013204,takamura2015estimating,davidov2010extraction} and comparative knowledge from the web \cite{tandon2014acquiring}. Our work uniquely learns verb-centric lexical entailment knowledge.

%
%

A handful of works have attempted to learn the types of knowledge we address in
this work. One recent work tried to directly predict several binary attributes
(such ``is large'' and ``is yellow'') from on off-the-shelf word embeddings,
noting that accuracy was very low \cite{rubinstein2015well}. Another line of
work addressed grounding verbs in the context of robotic tasks. One paper in
this line acquires verb meanings by observing state changes in the environment
\cite{she2016incremental}. 
Another work in
this line does a deep investigation of eleven verbs, modeling their physical
effect via annotated images along eighteen attributes \cite{gao2016physical}.
These works are encouraging investigations into multimodal groundings of a small
set of verbs. Our work instead grounds into a fixed set of attributes but
leverages language on a broader scale to learn about more verbs in more diverse
set of frames. In this, our work can be seen as exploring predicate lexical semantics in the vein of semantic proto-roles \cite{dowty1991thematic,kako2006thematic,reisinger2015semantic}, but instead affording pairwise, physical relations between roles .
\section{Conclusion}

\label{sec:conclusion}

We presented a novel take on verb-centric frame semantics to learn implied physical knowledge latent 
in verbs. 
Empirical results confirm that by modeling
changes in physical attributes entailed by verbs
together with
objects that exhibit these properties,
we are able to better infer new knowledge in both domains.

\section*{Acknowledgements}
This material is based upon work supported by the National Science Foundation Graduate Research Fellowship under Grant No. DGE-1256082, in part by DARPA CwC program through ARO (W911NF-15-1-0543), the NSF grant (IIS-1524371), and gifts by Google and Facebook. The authors thank the anonymous reviewers for their thorough and insightful comments. 

%

\bibliography{main}
\bibliographystyle{style/acl_natbib}

%

\end{document}